\documentclass[runningheads]{llncs}
\usepackage[T1]{fontenc}
\usepackage{graphicx}
\usepackage{amsmath}
\usepackage[linesnumbered,ruled]{algorithm2e}
\usepackage{cite}
\usepackage{setspace}
\usepackage{lineno} 
\usepackage{multirow}
\usepackage[colorlinks=true, urlcolor=blue, filecolor=green]{hyperref}
\usepackage{booktabs}
\usepackage{nicematrix}
\begin{document}
\title{FedCAR: Cross-client Adaptive Re-weighting for Generative Models in Federated Learning}

\titlerunning{FedCAR: New Aggregation Algorithm for Generative FL}
% If the paper title is too long for the running head, you can set 
% an abbreviated paper title here
%
\author{Minjun Kim\orcidID{0009-0008-1862-3966} \and Minjee Kim\orcidID{0000-0002-0035-4437} \and Jinhoon Jeong\orcidID{0000-0003-3255-1774}} 

\authorrunning{M. Kim et al.}
% First names are abbreviated in the running head.
% If there are more than two authors, 'et al.' is used.

\institute{Promedius Research, Seoul, South Korea\\
\email{\{mjkim1, mjkim, jhjeong\}@promedius.ai}}

\maketitle              % typeset the header of the contribution
\begin{abstract}
%The abstract should briefly summarize the contents of the paper in 150--250 words.
%TODO: local generator vs client generator 용어 선택 
Generative models trained on multi-institutional datasets can provide an enriched understanding through diverse data distributions. However, training the models on medical images is often challenging due to hospitals' reluctance to share data for privacy reasons. Federated learning(FL) has emerged as a privacy-preserving solution for training distributed datasets across data centers by aggregating model weights from multiple clients instead of sharing raw data. Previous research has explored the adaptation of FL to generative models, yet effective aggregation algorithms specifically tailored for generative models remain unexplored. We hereby propose a novel algorithm aimed at improving the performance of generative models within FL. Our approach adaptively re-weights the contribution of each client, resulting in well-trained shared parameters. In each round, the server side measures the distribution distance between fake images generated by clients instead of directly comparing the Fréchet Inception Distance per client, thereby enhancing efficiency of the learning. Experimental results on three public chest X-ray datasets show superior performance in medical image generation, outperforming both centralized learning and conventional FL algorithms. Our code is available at \href{https://github.com/danny0628/FedCAR}{https://github.com/danny0628/FedCAR}.
%\text{http://******.***/***}

\end{abstract}

\section{Introduction}
% FL과 PS에 대한 기본 설명
 Federated Learning (FL) is a machine learning framework designed to train models directly on clients' devices without data migration. In FL, a central server aggregates model weights from clients into a single global model, which is then shared back with the clients. This iterative process allows for collaborative model training while ensuring data privacy, as data remains on the local clients. The decentralized nature of FL offers a significant advantage in contexts where data security is crucial, such as in healthcare institutions, leading to its widespread use in medical image analysis \cite{Wicaksana2022FedMixMS, Wu2022FedRareFL, MRImageRecon, SplitAVG, Jiang2021HarmoFLHL}. Recently, generative adversarial network (GAN)\cite{Goodfellow2014GenerativeAN} in FL environment has drawn attention for their ability to handle a wide range of applications such as data anonymization\cite{Piacentino2024GeneratingF}, cross-domain adaptations \cite{Peng2020Federated} and other unsupervised learning tasks\cite{Das2022FGANFG, Ke2021StyleN}, while preserving privacy. \\ 
However, there are several obstacles in adapting FL to GAN. First, the implementation process can be tricky since it involves transmitting two distinct networks (the discriminator and generator) between the server and clients. Moreover, there are small reference code to implementation of federated GAN since previous researches about it is fairly limited. Second, previous studies\cite{ Li2022FederatedLG, math11194123, Zeng2022AdaptiveFLN} argued that non-independent and identically distributed data(non-i.i.d.) situations, which are commonly found in medical datasets, can cause instablity in convergence of training GAN as updates from different clients may be inconsistent. Thus, to address these obtacles, it is required to develop an advanced aggregation algorithm to embody federated GAN and train it effectively. \\
 To overcome these challenges, we propose a novel algorithm solely designed for generative models within FL frameworks. Our method includes adaptive re-weighting the contributions of client models based on quantitative evaluation of distribution discrepancies across clients. We utilized the Fréchet Inception Distance (FID) score to evaluate the distribution distances in feature representations across clients' data, directly addressing the heterogeneity in data distribution. We conducted experiments to show that this approach can significantly improve performance of generative model in non-i.i.d. settings. In addition, while conventional aggregation algorithms may tend to excessively overlook clients with smaller datasets, our approach avoids such issues by solely relying on clients' generative capabilities. \\
%\subsubsection{Related works}
Previous efforts to adapt FL to GAN have encountered significant hurdles, including poor performance \cite{Hardy2018MDGANMG} and limited applications to low-quality datasets \cite{Yonetani2019DecentralizedLO}. Recent studies have successfully demonstrated that FL-adapted GAN can achieve stable convergence by training the generator and discriminator in synchronization \cite{Rasouli2020FedGANFG, 10.1007/978-3-030-60636-7_1}.  
While the FedAvg \cite{McMahan2016CommunicationEfficientLO} aggregation method has been popular and effective in FL and chosen in FedGAN \cite{Rasouli2020FedGANFG}, there's a gap in exploring alternative aggregation methods. FedOpt \cite{reddi2021adaptive}, which leverages adaptive optimization, emerges as a potential advanced aggregation method option, but never been intensively studied. Building upon this foundation, our study aims to develop an effective aggregation algorithm for FL-adapted GAN by introducing a dynamic re-weighting of client contributions. \\
Our study have three contributions through this study. First, we developed a novel aggregation algorithm designed for generative models via cross-client evaluation and adaptive re-weighting in server-side, showing significant enhancement in chest x-ray image generation. Second, we show that generative models in FL can outperform centralized learning in non-i.i.d. settings. Finally, we release the source code for both the FL-integrated GAN and our adaptive re-weighting algorithm for FL (FedCAR), which has not been previously released.
% \ref{fig1}).
\begin{figure}[!t]
\label{fig1}
    \centering
    \includegraphics[width=1\linewidth]{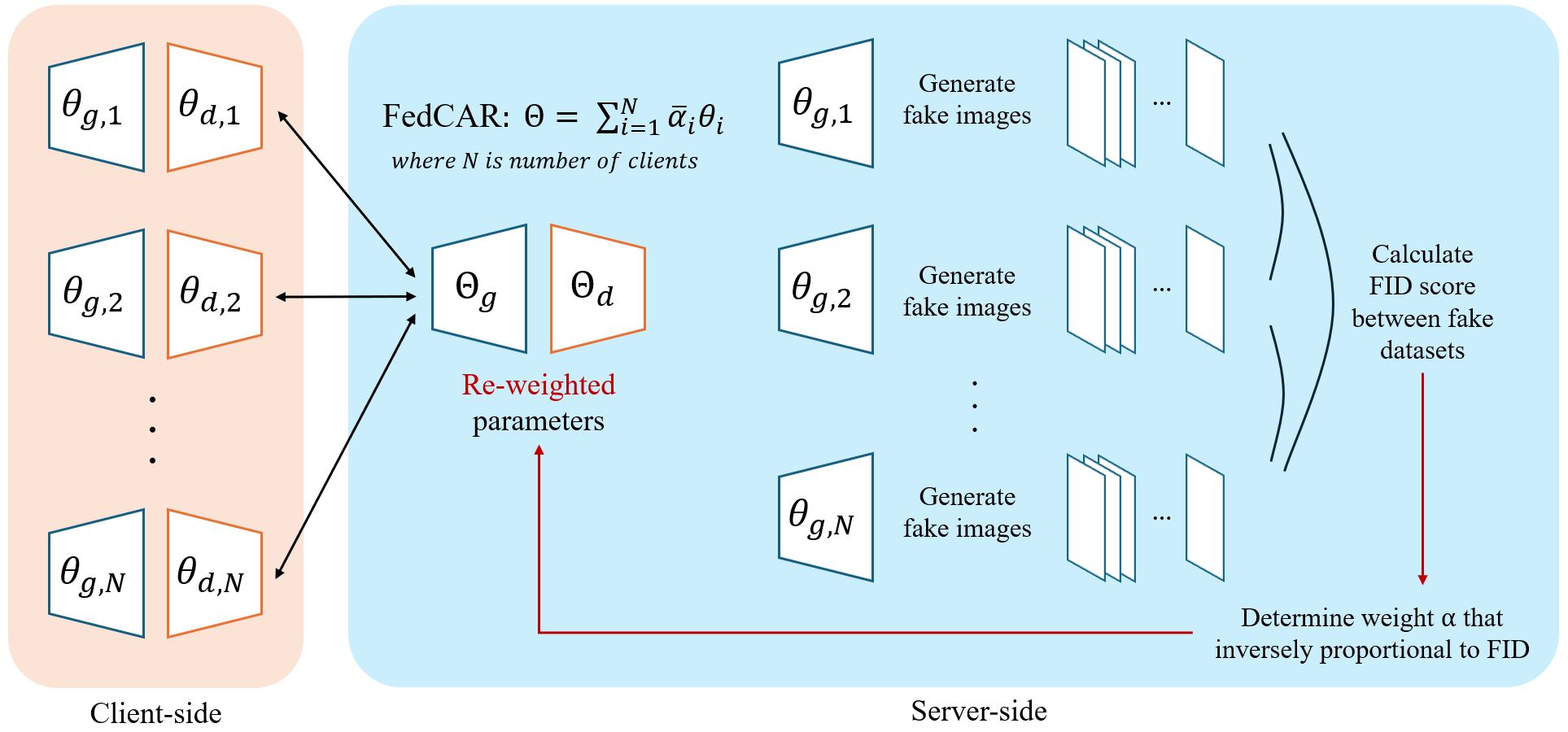}
    \caption{Overview of FedCAR aggregation algorithm}
    \label{fig:enter-label}
\end{figure}

\section{Method}

\subsection{Federated learning in GAN}
 In this study, we consider a real-world FL adaption for generative models, especially GAN, across $N$ clients that have own local dataset \(D_{i}\) where \(i \in \left\{1,2,...,N\right\}\). We designate $\theta$ to represent a GAN network that consists of the discriminator $\theta_{d}$ and the generator $\theta_{g}$. The essence of FL is collaboratively training a global model $\Theta$ by combining the model weights $\theta_{C}$ from ${N}$ clients, where $\theta_{C}$ represents the collective set of client models $\theta_{C}:=\{\theta_{i}\}_{i=1}^{N}$, in every training round $r$.\\
 As the references for our FL-related experiments, we train a centralized learning with whole datasets and three localized learning with individual dataset.
 %The parameter server collects $N$ sets of discriminators and generators, which are denoted as $\theta_{g, C}:=\{\theta_{g, i}\}_{i=1}^{N}$ and $\theta_{d, C}:=\{\theta_{d, i}\}_{i=1}^{N}$, respectively.
 % In our algorithm, the server assigns rewards or penalties to client models $\theta_{N}$ depending on the level of discrepancy among their distributions. Each generator $\theta_{g}$ produce synthetic datasets $X_N$ for FID-based comparison. 

\subsubsection{FedAvg in GAN}
Although FedAvg was not specifically designed for GAN, its intuitive and straightforward mechanism posed no implementation challenges for federated GAN. We simply followed its general process except aggregate twice for $\theta_{d}$ and $\theta_{g}$ respectively. Therefore, the global weight $\Theta$ can be described as below:
%Proof has been established in FedGAN that the two losses of GAN converge within the context of FL. Consequently, we have implemented GAN using FedAvg, the most representative algorithm in FL, this involves finding parameters that optimize the global loss and aggregating them as depicted by equation \ref{eqn1}.
% which is defined as a weighted average of the local loss functions \(L_{i}(w_{i})\) as below:
% \begin{equation}
%  L(W)= \sum_{i=1}^{N}\frac{m_{i}}{M}L_{i}(w_{i})
% \label{eqn1}
% \end{equation}
\begin{equation}
\label{eqn1}
%     % v^{k}_{t+1} = (g^{k}_{t} + d^{k}_{t}) - \eta_{t} \nabla F_{k}(\omega^{k}_{t}, \xi^{k}_{t})
%     T_k = \sum_{i=0}^{u} \nabla F_{k}(w_{t,i}) \;\; \; \Biggm/ \;\;\;w_{t+1, i \in u} = \sum_{i=0}^{u}(w_{t,i} -\eta \sum_{k=1}^{K} \frac{n_{k}}{n}w_{k,i}) 
% \label{eqn2}
\Theta = \frac{\sum_{i=1}^{N}d_{i}\theta_{i}}{\sum_{j=1}^{N}d_{j}} 
\end{equation}
where $d_{i}$ denotes the number of dataset belongs to $i$-th client.
\subsubsection{FedOpt in GAN}
FedOpt is a framework designed to improve FL by introducing optimization methodologies. To our knowledge, implementation of FedOpt to the generative model is not explored. Given that FedOpt employs adaptive optimization techniques at both the server and client level, it is anticipated that performance in the image generation will be enhanced as well. During local training stage, the client-side optimizer operates in each client as conventional optimizers. On the other hands, at the end of each round, the server-side optimizer operates by referencing the weights from the previous round. We prepared more detailed description of this process in Algorithm \ref{algo1}. 
%First, at the client-level, ClientOpt $x_{i,j+1}$ consist of $x_{i,j}$ of $r-1$ round received from server, after learning weight $g_{i,j}$, learning rate of the generator $\eta_{g}$, round information $r$. Also ClientOpt $y_{i,j+1}$ composed $y_{i,j}$ of $r-1$ round received from server, after learning weight $d_{i,j}$, discriminator learning rate $\eta_{g}$, round information $r$.
%These are grouped into a list in $\theta$ and sent to the server.
%At the server-level, FedAvg is performed with the received list $\theta$. ServerOpt $\theta_{r+1}$ is combined $\theta$, FedAvg value $\Delta_{r}$, server's learning rate $\eta$ and r as arguments.
%Afterwards, divide the results into $x_{r+1}$ and $y_{r+1}$ and enter them to complete the round. 
\begin{algorithm}[!h]
\label{algo1}
\caption{FedGAN adaptive OPT}
\DontPrintSemicolon
\SetKwInOut{Input}{Input}
    \Input{
        number of generator layer: $a$, number of discriminator : $b$ , \\
        ClientOPT, ServerOPT, $x_{0}$, $y_{0}$ \\
        
     }
  \For{$r$ = 0 to $R-1$}
    { 
        Set of clients $S = \{s_{1}, s_{2}, ..., s_{N}\}$ \\ 
        $\theta_{i,0}^{r}=\theta_{r}, x_{i,0}^{r}=x_{r}, y_{i,0}^{r}=y_{r}$ \\
        \For{each client i $\in$ S in parallel}
        {
            \For{$j$ = 0 to $J-1$}
            {
                Compute an unbiased estimate $g_{i,j}$ of $\nabla F_{i}(x_{i,j})$  \\
                Compute an unbiased estimate $d_{i,j}$ of $\nabla F_{i}(y_{i,j})$ \\ 
                $x_{i,j+1}$ = ClientOPT$(x_{i,j}, g_{i,j}, \eta_{g}, r)$ \\
                $y_{i,j+1}$ = ClientOPT$(y_{i,j}, d_{i,j}, \eta_{d}, r)$ \\
            }
        $\theta_{i,J}^{r}$ = $List(x_{i}^{r}[0]$ to $y_{i}^{r}[-1])$ \\
        $\Delta_{i}^{r}$ = $\theta_{i,J}^{r}$ - $\theta_{r}$ \\
        }
        $\Delta_{r}$ = $\frac{1}{\left\vert S \right\vert} \sum_{i \in S} \Delta_{i}^{r}$ \\
        $\theta_{r+1}$ = ServerOPT$(\theta_{r}, -\Delta_{r}, \eta, r)$ \\
        $x_{r+1}$ = $\theta_{r+1}[0:a]$ \\
        $y_{r+1}$ = $\theta_{r+1}[a:a+b-1]$ 
    }
\end{algorithm}
\subsection{Our Method : FedCAR}
\label{method:fedcar}
Presumably, during the training, several clients learned better than the others in every round. At the aggregation stage of each round, we intended to give more attention to the clients showed higher performance.
In our method, the server is responsible for evaluation of each client model's performance. By utilizing given generators from the clients, the server generates $x$ number of fake images and calculates FID score between them. We defined total FID of n-th client \(FID_{n}\) as,
\begin{equation}
FID_{n} = \sum_{i=1}^{n-1}FID_{i,n} + \sum_{j=1}^{N-n}FID_{n,n+j}
\label{eqn2}
\end{equation}
\noindent where \(N\) is total number of clients and \(FID_{c_{1},c_{2}}\) denotes FID score between fake dataset generated from local generator $c_{1}$ and $c_{2}$. Then we calculated parameter \(\alpha_{n}\) for each local models as inversely proportional to \(FID_{a,b}\). Finally, we can obtain the normalized \(\alpha\) which is required to re-weighted aggregation of local weights.
\begin{equation}
\alpha_{n} = \frac{1}{FID_{n}}
\label{eqn3}
\end{equation}
\begin{equation}
\overline{\alpha_{n}} = \frac{\alpha_{n}}{\sum_{i=1}^{N}\alpha_{i}}
\label{eqn4}
\end{equation}
$\overline{\alpha}$ is multiplied to each local weight which leads to pay more attention to better clients.
\begin{equation}
    % \Theta_{r+1} = \sum_{i=1}^{N}\overline{\alpha_{n}}\theta_{n}
    % \Theta_{r+1} = \sum_{i=1}^{N}\left({\alpha_{i}}\theta_{g_{i}} + \frac{\frac{n_{i}}{n}\theta_{d_{i}}}{N}\right)
    \Theta_{r+1} = \sum_{i=1}^{N}\alpha_{i}\theta_{g_{i}} + F\left(\theta_{d}\right)
\end{equation}
We performed above process at every round for dynamic and adaptive re-weighting. In Algorithm \ref{algo2}, we described the detailed process with pseudo code.

\begin{algorithm}[!ht]
\label{algo2}
\caption{FedCAR}
\DontPrintSemicolon
\SetKwInOut{Input}{Input}
    \Input{Global model $\Theta$, client model $\theta$, \\ 
     Number of clients $n$, number of kimg $k$, \\
     Number of rounds $r$}
  \For{$i$ = 0 to $r$}
    {
        $i$-th round start \\
        \For{$j$ = 0 to $k$}
        {$\theta^\prime \leftarrow $ Clients training ($\theta$)} 

        \For{$l$ = 0 to $n$}
        {
            Generate fake images: $X_l$ (Sample a subset) \\
            % $F_l \leftarrow \{FID(I_l, I_{l+1})\} \cup \{F_{n}=FID(I_{1},I_{n})\}$
            $F_{l \in n} \leftarrow \sum_{l=1}^{n} \sum_{m=1, m \ne l}^{n}FID(X_{l},X_{m})$
        } 
        
        $F_{bias} \leftarrow \frac{1}{F_{1}+F_{2}+ \cdots + F_{n} }$ \\
        $\Theta \leftarrow Aggregation(\theta^\prime \times F_{bias} ) $
    }
\end{algorithm}
\section{Experiments}
\subsection{Experimental Setup}
\subsubsection{Datasets}
Our study utilizes three public chest X-ray datasets: ChestX-Ray8 (NIH) \cite{8099852}, CheXpert \cite{Irvin2019CheXpertAL}, and VinDr-CXR \cite{Nguyen2020VinDrCXRAO}. We exclusively focus on normal images to clarify the comparison between aggregation algorithms in FL scenarios with non-i.i.d. datasets. All images were resized to 512 $\times$ 512 pixels and clipped the top 1\% percentile of pixel intensities to effectively reduced artifacts like L/R markers or devices as histogram enhancement. For experiment, we randomly sampled 10,000 images of "No finding" from each dataset, in total of 30,000 images. For our experiments, we randomly sampled 10,000 'No finding' images from each dataset, in total 30,000 images. For centralized learning, these were combined into a single dataset. In mild non-i.i.d. condition, each client was allocated 10,000 images from separate datasets. For severe non-i.i.d. condition, one client (CheXpert, chosen for its lower performance in individual learning) is trained using only 1,000 images, while the others have 10,000 images each, resulting in a total of 21,000 images. Considering different characteristics between datasets, we also conducted GAN training by clients individually.
\subsubsection{Implementation Details}
We utilized StyleGAN2\cite{Karras2019AnalyzingAI} as a baseline model, with a learning rate of $2e^{-3}$, Adam optimizer with betas as 0 and 0.99, and an exponential moving average of 10. We also exploited Flower framework \cite{Beutel2020FlowerAF} and the docker for the implementation of FL in real-world scenarios. We conducted experiments using 4 Nvidia-A100 80G GPUs: three allocated to clients and one served as the server. For transmission efficiency, we concatenated $\theta_{d}$ and $\theta_{g}$ into a single list for server communication (Algorithm \ref{algo1} line 11). At the server, concatenated lists from clients are aggregated into a global model $\Theta$ and send back to clients, where the clients subsequently split the list into $\theta_{d}$ and $\theta_{g}$ for resuming the local training (Algorithm \ref{algo1} line 16, 17). In our experiment, we set 100 for $k$ kimg per round and terminate rounds when it comes to the FID of all clients converges. During aggregation, FedCAR generates 1,000 fake images $x$ from each generators in center server, and measure FID scores between them as explained in section \ref{method:fedcar}

%\subsection{Comparison with State-of-the-art Methods}
\subsection{Result}
\begin{table}[ht]
\centering
\caption{Performance comparison result within the mild non-i.i.d. and the severe non-i.i.d. scenarios. In the mild case, the datasets have different characteristics, but amount of dataset for each client is the same. In the severe case, the amount of data belongs to CheXpert client is extremely less (10\%) than that of other clients.}
\setlength{\tabcolsep}{2.5pt}
\begin{tabular}{lcccccccc}
\toprule
& \multicolumn{4}{c}{Mild non-i.i.d.} & \multicolumn{4}{c}{Severe non-i.i.d.} \\
\cmidrule(r){2-5} \cmidrule(r){6-9}
Method & NIH & CheXpert & VIN & Avg.& NIH & CheXpert* & VIN & Avg. \\
\midrule
Centralized & - &  -& -& 8.58 & - & - & - & 12.99 \\ 
Individual & 11.87 & 12.76 & 5.49 & 10.04 & 11.87& 53.05 & \textbf{5.49} & 22.45 \\
\hline
FedAvg\cite{McMahan2016CommunicationEfficientLO}& 7.59 & 10.59 & 6.23 & 8.13 & 8.71  & 25.09 & 6.29  & 13.36\\
FedAdam \cite{reddi2021adaptive}& 8.13 & 10.54 & 6.18 & 8.28 & 11.42 & 45.43 & 8.27 & 21.70\\
\textbf{FedCAR(Ours)} &\textbf{6.62} & \textbf{9.80} & \textbf{4.94} & \textbf{7.12} & \textbf{8.28} & \textbf{22.99} & 7.02 & \textbf{12.76} \\
\bottomrule
\end{tabular}
%\bigskip
%\medskip
%* only 1,000 images 
\label{tab:my_label}
\footnotesize{$^*$10\% sized dataset}
\end{table}

\subsubsection{Mild Non-i.i.d. Scenario}
Let us assume that three institutions aim to collaboratively develop a generative model while ensuring no data transfer between them, to keep the data security. Each institution contributes a dataset containing 10,000 normal chest X-ray images, sourced from the NIH, CheXpert, and the VinDr datasets, respectively. Although the datasets are of comparable size, they exhibit unique characteristics attributable to their diverse origins. This scenario does not align precisely with the assumption of i.i.d. data, nor does it present extreme deviations from i.i.d. assumptions. Instead, it represents a scenario of mild non-i.i.d. characteristics, offering a realistic reflection of the complexities encountered in practical applications. \\
In this scenario, since re-weighting is performed cross-client adaptively in each round(fig \ref{fig:fedCAR}), our method preserved the characteristics of the client. As a result, FedCAR outperformed every other methods including centralized learning. In detail, FedCAR shows lower FID by 0.97, 0.74, 0.55 than the NIH, CheXpert, Vin dataset, respectively. It also show lower FID by 1.01 compare to the centralized learning. Table \ref{tab:my_label} summarized the detailed result.\\
Throughout the training process, $\alpha_{n}$ of clients exhibited minor fluctuations but generally remained around 0.33 (See Supplement Table 1). This suggests that our model achieved efficient learning by maintaining a balanced learning across all clients while incorporating slight variations per round.

\subsubsection{Severe Non-i.i.d. Scenario}
\begin{figure}[!t]
    \centering
    \includegraphics[width=1\linewidth]{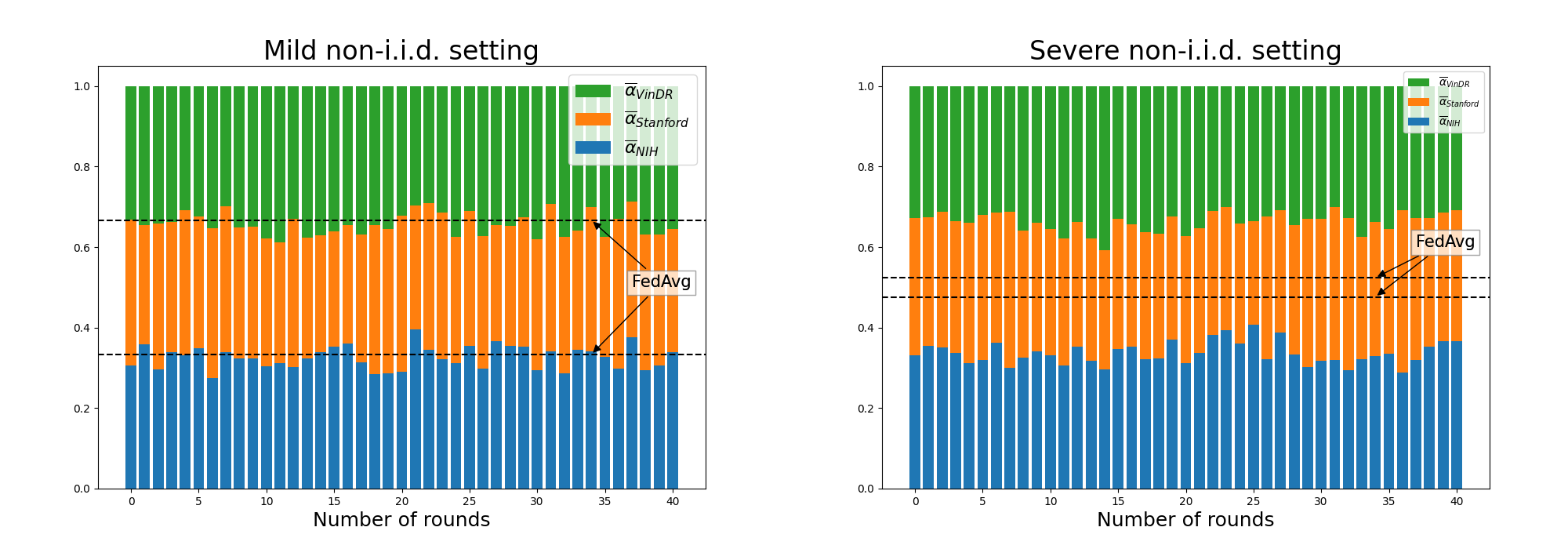}
    \caption{Variation of $\overline{\alpha_{n}}$ over training rounds.}
    \label{fig:fedCAR}
\end{figure}

We also explored more severe non-i.i.d. scenario that one of the institution has small amount of data compare to others. In this scenario, the institution would not dare to train generative model on their own since generally generative models consumes large amount of data. With FL, however, the institution may have chance to train generative models by other data-abundant institutions' assistance. Thus, we reduced number of CheXpert client from 10,000 to 1,000 images while other clients remain unchanged.
As a result, our method achieved the smallest FID score in average compare to other methods. FedCAR shows lower FID by 0.43, 2.10 than the NIH, CheXpert dataset, respectively. It also recorded lower FID by 0.23 compare to the centralized learning.\\
In addition, similar to the mild non-i.i.d. scenario, $\alpha_{n}$ hold ~0.33 during training rounds (See fig \ref{fig:fedCAR} and Supplement Table 1). Naturally, there is slight decrease in $\overline{\alpha}_{CheXpert}$ compare to previous scenario, due to deteriorated performance of the data-deficient client.

%%%
\section{Discussion}
Our algorithm, FedCAR, outperforms other aggregation algorithms in chest x-ray image generation regardless of discrepancies of data distributions among clients. While focusing on GAN in this study, the concept that utilization of server-side evaluation of the generator weight at each client can be broadly adapted to any other generative models in FL environments.\\
We observed that the averaged values of $\overline{\alpha_{n}}$ over training rounds is similar to each other in both scenarios. More specifically, $Average(\overline{\alpha_{n}})\approx0.33$ not only in the mild non-i.i.d. scenario but also the severe non-i.i.d. scenario, and this verifies that FedCAR does not re-weight by far compare to FedAvg in average(fig \ref{fig:fedCAR}). In the severe non-i.i.d. scenario, FedAvg reflects only 10\% of weights from CheXpert dataset as it has 10\% number of data compare to others. Instead, FedCAR improves learning efficiency by favoring clients with better performance in each training round. \\
We used the FID metric to evaluate performance of the algorithms. Although it does not directly measure perceptual quality and only shows similarity between the distributions, to our knowledge there is only one definitely better metric in image generation and it is evaluation by human experts. However, it requires much more cost and time so we will study with this evaluation in the future.\\
We are intrigued that federated GAN outperforms individual and centralized GAN in both scenarios. It is well recognized that FL under i.i.d. conditions is not inferior to centralized learning. However, the results of this experiment imply that, specifically for generative models, FL may be more efficient than centralized learning. We will further study about this issue to validate the hypothesis.

\section{Conclusion}
In our study, we discuss a novel FL aggregation algorithm designed for generative models, specifically addressing the challenges of training on multi-institutional medical image datasets while preserving privacy. By adaptively re-weighting clients' contributions to global model based on their grade in generating fake images, we enhance model training efficiency, image quality, and diversity. Experimental validation on public chest X-ray datasets confirms our method's superiority over both centralized learning and traditional FL approaches, showcasing its potential to advance medical image generation in a privacy-conscious manner.
\bibliographystyle{splncs04}
\bibliography{references}

\end{document}